\documentclass{article}

\usepackage{PRIMEarxiv}

\usepackage[utf8]{inputenc} 
\usepackage[T1]{fontenc}    
\usepackage{hyperref}       
\usepackage{url}            
\usepackage{booktabs}       
\usepackage{amsfonts}       
\usepackage{nicefrac}       
\usepackage{microtype}      
\usepackage{lipsum}
\usepackage{fancyhdr}       
\usepackage{graphicx}       
\graphicspath{{media/}}  
\usepackage{listings}
\usepackage[
backend=biber,
style=ieee,
sorting=none
]{biblatex}

\title{Aligning Large Language Models for Clinical Tasks
}

\author{
  Supun Manathunga \\
  University of Peradeniya \\
  NeuraSense Research \\
  \texttt{} \\
   \And
  Isuru Hettigoda \\
  NeuraSense Research \\
}

\begin{document}

\maketitle \

\begin{abstract}
Large Language Models (LLMs) have demonstrated remarkable adaptability, showcasing their capacity to excel in tasks for which they were not explicitly trained. However, despite their impressive natural language processing (NLP) capabilities, effective alignment of LLMs remains a crucial challenge when deploying them for specific clinical applications. The ability to generate responses with factually accurate content and to engage in non-trivial reasoning steps are crucial for the LLMs to be eligible for applications in clinical medicine. Employing a combination of techniques including instruction-tuning and in-prompt strategies like few-shot and chain-of-thought prompting has significantly enhanced the performance of LLMs. Our proposed alignment strategy for medical question-answering, known as 'expand-guess-refine', offers a parameter and data-efficient solution. A preliminary analysis of this method demonstrated outstanding performance, achieving a score of 70.63\% on a subset of questions sourced from the USMLE dataset.
\end{abstract}


\section{Introduction}

Until recent past, Artificial Intelligence (AI) research was mainly focusing on specific tasks like mastering the game of chess and Go \cite{campbell_deep_2002, noauthor_mastering_nodate-1}. However, the advancement of the deep learning techniques, particularly the transformer models has revolutionized the way humans interact with AI models, especially in the realm of Natural Language Processing (NLP) \cite{vaswani_attention_2023-1}. The transformer architecture has laid the groundwork for Large Language Models (LLMs), which exhibit the remarkable capacity to perform tasks they weren't explicitly trained for, a phenomenon observed as these models are scaled to substantial capacities \cite{bommasani_opportunities_2022}. The development of these expansive LLMs may be  bringing us closer to the threshold of realizing Artificial General Intelligence \cite{goertzel_artificial_2014, bubeck_sparks_2023-1}. 

LLMs have been trained on large text corpora containing medical knowledge and this knowledge becomes ingrained in their neural weights \cite{singhal_large_2022-3}. Capitalizing on the task-agnostic nature of LLMs, they find utility across a spectrum of clinical medicine tasks, ranging from information retrieval and summarization to decision-making and diagnostics \cite{bommasani_opportunities_2022}. However, given the sensitive nature of clinical medicine, it is imperative for these models to aptly grasp the nuances of tasks, extract pertinent information, and engage in reasoned analysis with a certain level of discernment. Mechanisms have to  be devised to mitigate hallucination, guarding against harmful content, and ensuring the model's alignment with medical ethics \cite{lievin_can_2023-1, bender_dangers_2021, noauthor_considering_nodate}.

Therefore, despite impressive NLP capabilities exhibited by LLMs, they need to be aligned before deploying for specific clinical tasks \cite{singhal_towards_2023-1}. Different alignment techniques try to achieve different goals. Instruction finetuning or instruction-tuning trains the model to follow human instruction better, making the model outputs more truthful, less toxic, and structured in a specific way \cite{ouyang_training_2022}. Finetuning has been one of the most utilized methods, which involves adjusting the weights of the pre-trained model via a supervised dataset, typically with thousand to hundreds of thousands of examples \cite{brown_language_2020-1}.  The disadvantages in this approach are the need of a fresh dataset tailored to each new task and the compute-heavy nature of the process \cite{brown_language_2020-1}. 

An alternative for this is few-shot learning, wherein the model is expected to perform a specific task with only a handful of demonstrations prepended to the input context \cite{liu_pre-train_2021-3}. Chain-of-thought (CoT) prompting is another technique which improves the model’s reasoning prowess inducing a step-by-step thinking process akin to human cognition \cite{wei_chain--thought_2023-1}. In-context prompting strategies like few-shot prompting and chain-of-thought (CoT) have led to substantial enhancements in reasoning capabilities, obviating the need for task-specific datasets. However, the performance of these approaches might not match that of finetuned models \cite{wei_chain--thought_2023-1, brown_language_2020-1}.

Studies have explored the effectiveness of training smaller-scale LLMs exclusively on scientific and biomedical corpora \cite{beltagy_scibert_2019-1, lewis_pretrained_2020-1, shin_biomegatron_2020-1, noauthor_biobert_nodate-1, gu_domain-specific_2022-1, hong_diminishing_2023-1,noauthor_biogpt_nodate-1}. Given the multifaceted nature of diverse clinical tasks, it is reasonable to anticipate that larger models with heightened reasoning capacities would outperform smaller counterparts, particularly when complemented by refined alignment methodologies, as opposed to smaller models trained on meticulously curated datasets \cite{singhal_large_2022-3,lievin_can_2023-1,singhal_towards_2023-1}. Yet, the recently released PubMedGPT 2.7B model challenged this notion by achieving a score of 50.3\% on the USMLE dataset \cite{noauthor_stanford_nodate-1}.

\section{LLMs as medical question-answering models}
\label{sec:headings}

The USMLE dataset, which is a subset extracted from the larger MedQA dataset, comprises multiple-choice questions sourced from the medical board exams in the United States \cite{jin_what_2020-1}. Typically, these questions demand multi-hop reasoning that traverses a spectrum of medical knowledge. They are commonly employed alongside other datasets to evaluate and benchmark the performance of Large Language Models (LLMs) in the context of medical question answering \cite{noauthor_papers_nodate}.

Several studies have investigated the utility of large-scale LLMs in medical question answering \cite{singhal_large_2022-3,lievin_can_2023-1,singhal_towards_2023-1}. Liévin et al. found that the code-finetuned code-davinci-002 175 B parameter GPT-3.5 series model scored 53.1\% when was combined with retrieval augmentation and multiple-prompting on USMLE dataset \cite{lievin_can_2023-1,jin_what_2020-1}. They have used a BM25 retriever made of Wikipedia articles for grounding \cite{manning_introduction_2009-1}. The study has shown that even without using retrieval augmentation, zero-shot GPT-3.5 performance was superior to that of finetuned BERT, indicating that GPT-3.5 was able to leverage implicit knowledge and reasoning better in the domain of USMLE question-answering tasks. Therefore, the researchers inferred that LLMs of the scale of GPT-3.5 family can efficiently tap into the parametric medical knowledgebase and execute non-trivial reasoning steps. Furthermore, the study demonstrated that when the inference-time compute was sufficiently increased by sampling multiple generations through CoT, such models can virtually surpass the pass-mark for USMLE \cite{lievin_can_2023-1}. 

With increasing utilization of LLMs, the necessity for comprehensive benchmarks to evaluate them across different domains emerged. Singhal et al. aggregated various existing medical question-answering (QA) datasets with the addition of a new dataset that encompasses commonly searched health questions to curate the dataset MultiMedQA \cite{singhal_large_2022-3,jin_what_2020-1,hendrycks_measuring_2021-3,pal_medmcqa_2022-1,jin_pubmedqa_2019-1}. The authors have also developed an instruction prompt tuning technique which is both data and parameter efficient in aligning LLMs to medical domain tasks. Their model built upon an instruction-tuned variant of the 540 B parameter PaLM model (Flan-PaLM) exhibited exceptional performance on the USMLE dataset with an accuracy of 67.6\% \cite{chowdhery_palm_2022-3}. This achievement was made possible through a combination of strategies including few-shot prompting, chain-of-thought and self-consistency \cite{wang_self-consistency_2023}.

In April 2023, Microsoft and OpenAI published their results of GPT-4 on medical benchmarks \cite{nori_capabilities_nodate-1}. GPT-4 may be the largest language model ever created even though OpenAI has not disclosed the exact number of parameters in the model. Some experts speculate that its parameter count exceeds 1.7 trillion \cite{schreiner_gpt-4_2023}. The GPT-4 base model without any finetuning scored 83.76\% on USMLE dataset on zero-shot prompting \cite{nori_capabilities_nodate-1}. This accomplishment potentially underscores efficient knowledge retrieval and advanced reasoning with increasing model size and training data \cite{kaplan_scaling_2020}.

Google Research and DeepMind announced the model Med-PaLM 2 in May 2023, as an improvement over their preceding iteration, Med-PaLM \cite{singhal_towards_2023-1}. They have used medical domain-specific finetuning and a novel prompting strategy termed ensemble refinement. The ensemble refinement technique draws from a foundation of chain-of-thought prompting, self-consistency, and self-refinement mechanisms \cite{madaan_self-refine_2023}. This two-step process begins by sampling multiple generations, each accompanied by explanations via few-shot Chain-of-Thought (CoT) prompting. The second step involves combining the initial question with the concatenated multiple generations from the previous stage, and generating a refined answer. The model achieved state-of-the-art performance on the USMLE dataset with an accuracy of 85.4\%.

\subsection{Mitigating factual inconsistency}

Prior research has investigated the dual-role of LLMs as implicit knowledgebases and reasoning models \cite{hendrycks_measuring_2021-3,taylor_galactica_2022-3,joshi_triviaqa_2017-5}. The parameterized knowledgebase encoded in the model weights cannot be easily updated or expanded. It is difficult to ‘prove’ the factual accuracy of generated responses because of the implicit nature of the knowledgebase, which functions as a latent representation of the training data \cite{lewis_retrieval-augmented_2021, guu_realm_2020-3}. Factual inconsistencies and the potential for generating inaccurate information present significant obstacles when leveraging the LLM's latent knowledgebase, especially in sensitive domains like medicine \cite{huang_factual_2023}. Additional mechanisms need to be implemented to verify the outputs generated by LLMs in such occasions \cite{manakul_selfcheckgpt_2023}. 

Integrating a non-parametric memory with the LLM to create a hybrid model offers a promising solution to address some of these challenges. Defining an explicit knowledgebase and augmenting the LLM generation with the retrieved information from the non-parametric memory makes it possible to examine the source of the information of the LLM generated output \cite{lewis_retrieval-augmented_2021}. Several studies have examined the performances of such Retrieval Augmented Generation (RAG) models when both the retriever and the generator were trained end-to-end \cite{lewis_retrieval-augmented_2021,guu_realm_2020-3}. These studies have showcased superior performance on open-domain question-answering benchmarks \cite{joshi_triviaqa_2017-5, kwiatkowski_natural_2019-1,berant_semantic_2013,baudis_modeling_2015-1}.

\subsection{Explainable knowledge and reasoning}

It is evident that with progressive upscaling, finetuning and improved prompting strategies, LLMs are acquiring the ability to manipulate clinical knowledge. Nevertheless, it becomes imperative to employ transparent mechanisms for knowledge retrieval and reasoning, aligning with the demands of clinical medicine where the precision of information holds paramount importance \cite{manathunga_retrieval_2023,reddy_evaluating_2023-2}. In this context, the utilization of an explicit non-parametric knowledgebase gains significance. Such knowledgebases can be easily updated and are data and parameter efficient since the LLM does not need to be retrained to infuse new knowledge \cite{verga_adaptable_2021}. 

We observed the vulnerability of LLMs to diversion into incorrect lines of reasoning, potentially due to the undue emphasis placed on irrelevant contextual information, leading to the generation of unrelated or potentially harmful outputs \cite{shi_large_2023}. The common benchmarks that are used to evaluate the performance of LLMs in clinical settings including USMLE mostly comprise relevant information to arrive at the correct answer. However, real-world instances frequently encompass extraneous information that necessitates the model's ability to discern and discount such distractions. It has been shown that when irrelevant information appears in the context, LLMs tend to make mistakes unless specific measures like instructed prompting and introduction of exemplar challenges containing distractors are implemented \cite{shi_large_2023}.

In an attempt to overcome these problems, we propose a strategy involving retrieval augmented generation using dense vectors and a prompting strategy termed ‘expand-guess-refine’. This strategy operates in a zero-shot manner, without model finetuning, rendering it considerably computationally efficient than preceding methods.

\section{Methodology}
\label{sec:others}

\subsection{Model}

The LLM that was used for the preliminary evaluation was OpenAI \verb|gpt-3.5-turbo| 175B parameter model.

\subsection{Vector database}

The vector database was compiled by segmenting the text of 18 medical books, which were originally collected as PDF versions and converted into text via optical character recognition. The books were released upon the license agreement of research use only, with the MedQA dataset. There were 231,581 total paragraphs containing 12,727,711 tokens. Based on a preliminary analysis conducted by Jin et al., human experts could find enough evidence 88\% of the times to answer a random set of 100 questions from the development split of the USMLE dataset. However, only 2\% of the questions assessed the knowledge on a single knowledge point while the rest of the questions simulated complex clinical cases \cite{jin_what_2020-1}.

The whole text corpus of all the books was split into chunks of maximum of 3000 characters with 1000-character overlaps using recursive text splitter (RTS) method. RTS tries to split the text based on a parameterized list of characters. The splits were subsequently embedded using 1536-dimensional OpenAI text-embedding-ada-002 embedding model and stored in FAISS vector database \cite{johnson_billion-scale_2017-2}.

\subsection{Expand-guess-refine prompting}

This prompting strategy consists of three components.

\subsubsection{Expand}

We observed that in some cases where the context of a question, and potentially even the question itself, is presented in a concise manner, there is a tendency for LLMs to overlook critical contextual elements or to miscomprehend the question.

The Expand strategy reshapes the context by expanding it and elaborating on important points. This is followed by a rephrasing of the question into a direct query format, rather than being presented as a multiple-choice question (MCQ). This step does not involve the non-parametric knowledgebase.

The exact prompt that is used is

\begin{lstlisting}

    You will be given a context and a multiple-choice-question at the end. Identify the context and the question separately. Your task is to expand the passage and the question fully to make it easy to understand, and to think step bystep. Rephrase the question as a direct question, not as an MCQ.

{question}

Expanded context:

Direct question:
\end{lstlisting}

\begin{figure}
    \centering
    \includegraphics[width=01\columnwidth]{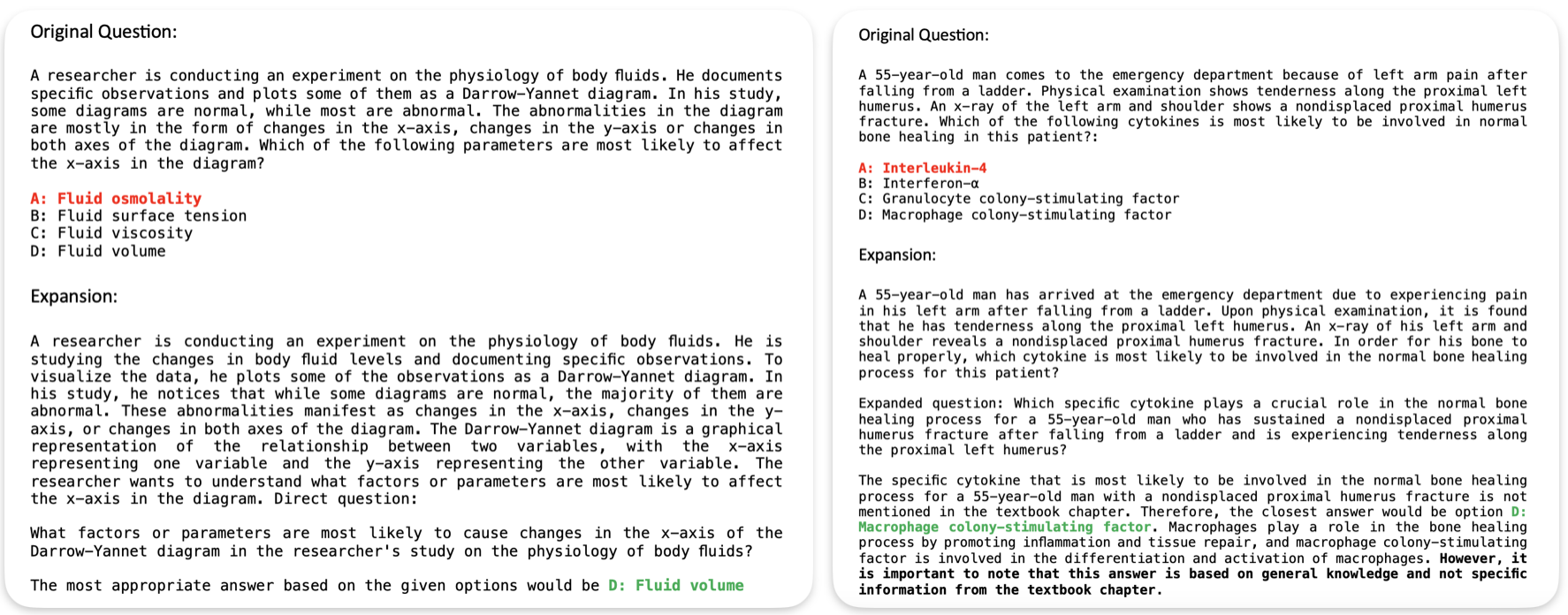}
    \caption{Expansion of the context.}
    \label{fig:Figure 1}
\end{figure}

Figure \ref{fig:Figure 1} is an example of the model deriving the correct answer only using the Expand strategy, without utilizing the non-parametric knowledgebase.

\subsubsection{Guess}

The observation has been made that the mere presence of a single unrelated word within the context can steer the LLM towards generating an entirely erroneous answer. This propensity is particularly pronounced in the context of Multiple-Choice Questions (MCQs), where only one option is correct. In light of this, the Guess stage is introduced, aiming to predict the response to the expanded question before seeing the answer choices, with the assistance of top-k retrieved documents sourced from the vector database.

The prompt is as follows.

\begin{lstlisting}
Read the following passage and the question.

{question}

Now read the TEXTBOOK CHAPTER.

BEGIN TEXTBOOK CHAPTER
 
{context}

END OF TEXTBOOK CHAPTER

Provide the answer to the question in the passage, with the help of details from the TEXTBOOK CHAPTER

\end{lstlisting}

\begin{figure}
    \centering
    \includegraphics[width=01\columnwidth]{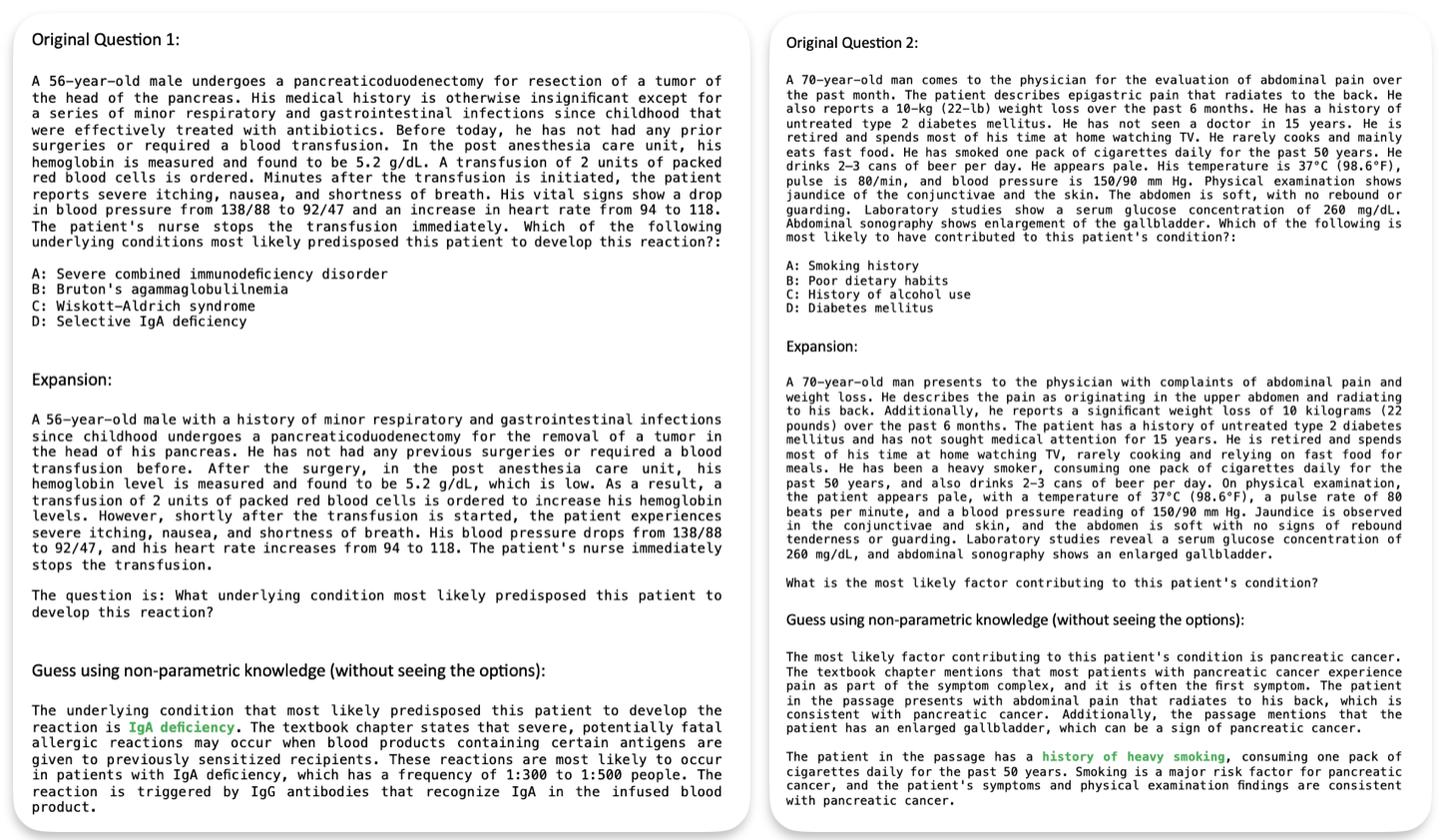}
    \caption{The Guess strategy.}
    \label{fig:Figure 2}
\end{figure}

Figure \ref{fig:Figure 2} is an example of the use of Guess prompt.

\subsubsection{Refine}

The final stage compiles a prompt using the transformed context, the guess and the actual options provided. The LLM is tasked with selecting the most appropriate answer from the provided answer choices. This selection process draws upon the generated guess as well as the relevant documents retrieved from the vector database, collectively guiding the model's decision-making. The prompt template is

\begin{lstlisting}
    {Expansion}

    {Guess}
        

    THINK STEP BY STEP. NOW SELECT THE BEST ANSWER FOR THE QUESTION OUT OF THESE FOUR OPTIONS.IF THE MOST APPROPRIATE ANSWER IS NOT THERE, SELECT THE CLOSEST ANSWER. YOU MUST SELECT AN ANSWER FROM THE OPTIONS. GIVE REASONS FOR CHOOSING THAT ANSWER. IF THE TEXTBOOK CHAPTER DOES NOT CONTAIN THE ANSWER, GIVE THE ANSWER BASED ON YOUR KNOWLEDGE.

    Options:

    {Options}
\end{lstlisting}

\begin{figure}
    \centering
    \includegraphics[width=0.7\columnwidth]{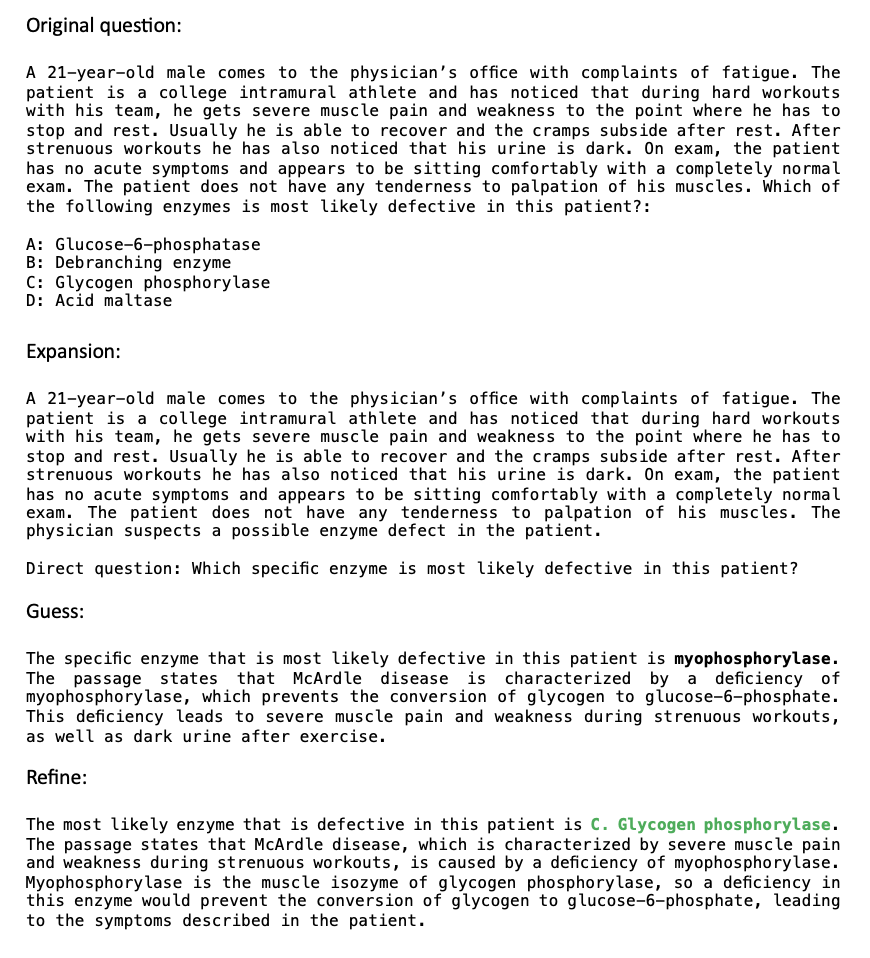}
    \caption{Expand-guess-refine strategy}
    \label{fig:Figure 3}
\end{figure}

An example is shown in Figure \ref{fig:Figure 3}.

\section{Preliminary Analysis}

A preliminary analysis was conducted on the first 100 questions and 50 random questions from the USMLE development data split. Seven image-based questions were excluded and the model achieved an accuracy of 70.63\% while the accuracy achieved by chatGPT was 59.44\%. The improvement achieved by the expand-guess-refine model was statistically significant (p-value 0.031) for two-sample test for equality of proportions. 

\section{Discussion}

With the recent advancements in LLMs, their integration within healthcare has been evaluated in a diverse set of tasks such as summarizing patients’ health records, writing discharge summaries, getting assistance for medical research and evaluating clinical scenarios to formulate differential diagnoses \cite{jin_what_2020-1,reddy_evaluating_2023-2, arora_promise_2023-1,yang_gatortron_2022-1}. In our perspective, the eligibility of LLMs for deployment in clinical related tasks hinges on their proficiency in two critical dimensions: their capability to serve as a robust and reliable knowledge repository and their capacity to effectively function as intelligent processors of natural language.

Medicine is a rapidly evolving field. More than 1.3 million new citations have been indexed in MEDLINE database in the fiscal year 2022 \cite{noauthor_citations_nodate-1}. Therefore, it is important to utilize methods that facilitate the seamless updating of LLM knowledgebases. It is equally important that the knowledge to be explainable. The factual accuracy of the generations of the LLM should be verifiable by inspecting the sources of information of the LLM generated content. The preliminary analysis of this study has suggested that augmenting the implicit knowledgebase of the LLM with a high-quality task-specific non-parametric knowledgebase can significantly improve performance as well. 

LLMs create an internal latent representation of the training data which is pivotal for achieving generalization. Consequently, the outputs generated by LLMs can exhibit a form of ‘apparent’ reasoning \cite{sejnowski_large_2023-1}. However, it is essential to recognize that this type of reasoning or 'thought process' significantly diverges from human cognitive processes. In evaluating the logic underpinning the LLM generated content, there arises a need to translate the apparent reasoning of LLMs into a step-by-step framework akin to human thinking. This translation is fundamental for gauging the coherence and accuracy of the generated logic. Thus, the dependability of LLM-generated content depends not solely on the capability to produce intelligible knowledge but also on the capacity to generate comprehensible reasoning.

In addition to model finetuning, aggregating results across multiple generations and in-prompt alignment strategies like few-show CoT, we have demonstrated in this preliminary analysis that retrieval augmentation and expand-guess-refine prompting can significantly improve LLM performance with the additional advantages of generating explainable knowledge and reasoning.

Code is available at \href{https://github.com/ssm123ssm/medGPT}{https://github.com/ssm123ssm/medGPT}

\medskip

\printbibliography

\end{document}